\documentclass[10pt,twocolumn,letterpaper]{article}

\usepackage{cvpr}
\usepackage{times}
\usepackage{epsfig}
\usepackage{graphicx}
\usepackage{amsmath}
\usepackage{amssymb}
\usepackage[breaklinks=true,bookmarks=false]{hyperref}

\cvprfinalcopy 

\def\cvprPaperID{****} 

\begin{document}

\title{Compliance Checking with NLI: Privacy Policies vs. Regulations}

\author{Amin Rabinia\\
University of Maine\\
Orono, ME\\
{\tt\small amin.rabinia@maine.edu}
\and
Zane Nygaard\\
University of Maine\\
Orono, ME\\
{\tt\small zane.nygaard@maine.edu}
}

\maketitle

\begin{abstract}
  
A privacy policy is a document that states how a company intends to handle and manage their customers’ personal data. One of the problems that arises with these privacy policies is that their content might violate data privacy regulations. Because of the enormous number of privacy policies that exist, the only realistic way to check for legal inconsistencies in all of them is through an automated method. In this work, we use Natural Language Inference (NLI) techniques to compare privacy regulations against sections of privacy policies from a selection of large companies. Our NLI model uses pre-trained embeddings, along with BiLSTM in its attention mechanism. We tried two versions of our model: one that was trained on the Stanford Natural Language Inference (SNLI) and the second on the Multi-Genre Natural Language Inference (MNLI) dataset. We found that our test accuracy was higher on our model trained on the SNLI, but when actually doing NLI tasks on real world privacy policies, the model trained on MNLI generalized and performed much better.
  
\end{abstract}

\section{Introduction}

When companies collect personal data of customers, they are required to publish a privacy policy to state how they intend to handle the personal data. This privacy policy needs to inform users what kind of data will be collected and how their data will be managed. One example where people frequently come into contact with a privacy policy is when they install an app on their phone. Unfortunately, these privacy policies are usually very lengthy and, because the language is legal in nature, they are difficult to read for a layperson. Due to such difficulties, when users are presented with the privacy policy and they have to choose whether to agree to it, most users never read the policy and simply accept it. However, we have very little oversight for these privacy policies and whether they are actually compliant with the privacy regulations.

Automating the compliance checking of the privacy policies is an appealing solution. This task can be done with helps from Natural Language Inference (NLI) techniques. For any NLI task there is a pair of premise and hypothesis, and a label that mentions the relation between the premise and hypothesis. Given a pair of premise and hypothesis, a model must be able to decide on whether the premise entails the hypothesis, whether it contradicts the hypothesis, or whether it is neutral. The model will then assign the label accordingly. This task is also called Recognizing Textual Entailment or RTE. Here is an example of text entailment recognition:
\begin{quotation}
\noindent
    \textbf{Premise:} A black race car starts up in front of a crowd of people.\\
    \textbf{Hypothesis:} A man is driving down a lonely road.\\
    \textbf{Label:} contradiction 
\end{quotation}
In this case, we would have a contradiction because the sentence stated in the hypothesis opposes the premise.

For our NLI task we use two different models with each model using a different dataset. The first one is Stanford Natural Language Inference (SNLI) dataset~\cite{bowman2015large}. The SNLI is a dataset that consists of a ``collection of 570k human-written English sentence pairs manually labeled for balanced classification with the labels entailment, contradiction, and neutral.'' To create this dataset they had people look at a caption of a photo, this caption forming the premise for the NLI, and then write sentences about the caption with one being definitely true, one being possibly true, and one being definitely false. The sentences they wrote formed the hypotheses for this dataset. Therefore, the data samples in this dataset have a descriptive language. 

The second dataset we use is the MNLI~\cite{nangia2017repeval}. This dataset consists of “433k sentence pairs annotated with textual entailment information” and is seen as a successor to the SNLI dataset. Unlike the SNLI, this dataset is not based on the descriptions of pictures. This dataset has sentence pairs from written text, spoken text, and a wide range of other genres. Because of the wider range of genres, this dataset allows models trained on it to generalize far better than one might find with the SNLI.

Among the valuable work in natural language inference, Liu~\etal~\cite{liu2016learning} and Chen~\etal~\cite{chen2016enhanced} have proposed different approaches to solving the challenge of NLI. Liu~\etal were one of the first users of an inner-attention mechanism in their model, which they found a lot of success with and can now be seen in many modern NLP models. Whereas Chen~\etal aimed to improve existing methods of sequential LSTM chains by making some additions like syntactic parsing trees. With each of their respective approaches they were able to achieve state-of-the-art results.

Among the work for automating compliance checking and helping users with their privacy preferences, Story~\etal~\cite{story2019natural} proposes a classification approach for privacy policies. They created a model called the Mobile Apps Privacy System or MAPS that specifically looks at privacy policy compliance for apps on the Google Play Store. Through their automated approach they were able to look at compliance for over a million apps on the Google Play Store and they found that possible compliance issues were widespread. This shows how the lack of oversight for these privacy policies can be a real problem. This issue of little oversight and privacy regulation non-compliance are what we tackle in our work. The MAPS approach is one that uses NLP to look at a privacy policy for a particular app. They then use static analysis in a containerized environment where they are actually looking at what the app is doing at a software level and whether that matches up to what the privacy policy says it will do. What we add with our approach is that we are solving this problem through a purely NLI solution, without any kind of static analysis approach like MAPS. We are not looking at how an app, for example, actually runs and collects information and we are not focused specifically on an app running in a software environment. Our approach is purely using natural language inference and looks at whether there are discrepancies between what the privacy regulations are and what a company’s privacy policy says. Our approach could potentially handle a wider range of policy checking without needing a specific environment with an app in order to check for compliance. Our approach can be used, for instance, on a privacy policy put out by a company that states how they handle data that is collected through their website. 

\section{Related Work}

Previous work on improving NLI was done by Liu \etal~\cite{liu2016learning} using bidirectional LSTMs and self-attention approaches they achieved former state-of-the-art results of 85.0\% test accuracy on the SNLI (Stanford Natural Language Inference) dataset. Prior to their paper, the state-of-the-art was utilizing a simple attention-based mechanism rather than the inner-attention approach they used. Chen \etal~\cite{chen2016enhanced} aimed to improve the model of a sequential chain LSTM using their ESIM (Enhanced Sequential Inference Model). They showed that using a simple chain LSTM model rather than a complicated network can be quite effective when carefully designed. They first showed that using their basic ESIM approach they were able to beat the previous state-of-the-art results. Using recursive architectures and syntactic tree parsing they were then able to get their best result of 88.6\% test accuracy, which were former state-of-the-art results on SNLI.

More recently, Wang~\etal~\cite{wang2018glue} and Mitra~\etal~\cite{mitra2019understanding}, aiming to improve natural language problem solving, created robust datasets or benchmarks that can help models to generalize better than something that was trained only on a dataset like the SNLI. The paper by Mitra~\etal aims to solve the issue of models not handling the idea of “roles and entities” particularly well when trained on a limited dataset. They created two datasets with the aim of resolving this issue. The other paper related to broadening out the NLP infrastructure was by Wang~\etal and they were more broad in their scope, not focusing specifically on NLI. The benchmark they created, called the General Language Understanding Evaluation benchmark or GLUE~\cite{wang2018glue}, utilizes multiple existing reputable datasets with varying NLP tasks. At present, GLUE is used frequently as a difficult benchmark to test a model on. Many models struggle to handle the wide range of tasks and genres that are present in the benchmark. Another recent paper we have from 2019 is authored by Liu~\etal~\cite{liu2019roberta} and presents their variation on the BERT model, called RoBERTa. With this approach they make various changes to the BERT model including longer training with a larger dataset and working with longer sequences. Overall they were able to achieve state-of-the-art results on three benchmarks: GLUE, RACE~\cite{lai2017race}, and SQuaD~\cite{rajpurkar2016squad}. They also presented a new dataset called CC-NEWS that contained millions of news articles.

More recently in 2019 Wang \etal~\cite{wang2018glue} created a NLU (Natural Language Understanding) benchmark called GLUE (General Language Understanding Evaluation). Their approach is intended to resolve the issue of models being trained on a very limited type of problem or genre and not being able to generalize effectively. For this approach they used pre-existing datasets such as SNLI, MultiNLI, and QNLI. Their reasoning for using existing datasets is that these datasets are agreed upon and well-tested methods used by NLU researchers. Mitra \etal~\cite{mitra2019understanding} created two datasets to tackle the same problem that Wang \etal were trying to solve with models not generalizing well when they are only trained on a dataset like SNLI. In this paper, their approach is to develop datasets that try to help models understand roles and entities, which is something they observed many models struggle with when the models are failing to generalize. An example the authors give is that the ESIM model predicts entailment with 96.29\% confidence for the premise “Kendall lent Peyton a bicycle” and the hypothesis “Peyton lent Kendall a bicycle”. They also propose a new attention function that performs just as well on existing benchmarks as models they tested that are not modified with this new attention function like ESIM and DecAtt. The model they propose also performs better on the datasets they developed for this study than the other models they tested.

Also in 2019 we have a BERT (Bidirectional Encoder Representations from Transformers) pretraining approach taken by Liu \etal~\cite{liu2019roberta} called RoBERTa (A Robustly Optimized BERT Pretraining Approach) shows that “BERT was significantly undertrained, and can match or exceed the performance of every model published after it.” The modifications the authors make for RoBERTa include “(1) training the model longer, with bigger batches, over more data; (2) removing the next sentence prediction objective; (3) training on longer sequences; and (4) dynamically changing the masking pattern applied to the training data”. They also collected data for a new large dataset, CC-NEWS, for use in training, which contains millions of news articles. After pre-training using the BERT model with the new modifications (RoBERTa), it was evaluated on three different benchmarks, GLUE, SQuaD (Stanford Question Answering Dataset), and RACE (Reading Comprehension from Examinations). State-of-the-art results were achieved on the three benchmarks.

A paper from 2019 that is particularly similar to our work is from Story \etal~\cite{story2019natural}. They used an NLP approach called MAPS (Mobile App Privacy System) to analyze 1,035,853 apps on the Google Play Store for privacy policy compliance with the regulations. They found ``Potential compliance issues appeared to be widespread, and those involving third parties were particularly common.'' To put this method into practice they took a three-tiered approach: ``a privacy  practice  statement  is  classified  based  on  a  datatype (e.g., location), party (i.e., first or third party), and modality (i.e., whether a practice is explicitly described as being performed or not performed).'' They then used a ``pipeline of distributed tasks  implemented in a containerized software stack.'' By using this containerized setup they performed static analysis and were able to see whether the actual actions that the app was taking was following what the privacy policy stated the app would do. They found ``many apps' privacy policies do not sufficiently disclose identifier and location data access practices performed by ad networks and other third parties.'' Using this method they had negative F1 scores ranging from 78\% to 100\%, which they say suggests an improvement over the state-of-the-art.

\section{Design}

In this work, we tried two different network designs for the SNLI and MNLI datasets. While Design-I was satisfactory on SNLI, it fell short in handling the content diversity of MNLI. Therefore, we designed a model with more layers for training on MNLI (Design-II). 

For word vector embedding we used GloVe 6B 300D, a pretrained embedding layer with 400k unique words~\cite{pennington-etal-2014-glove}. 

\subsection{Design-I for SNLI}
Stanford Natural Language Inference (SNLI)~\cite{bowman2015large} corpus is the biggest existing dataset for NLI, which contains 570k pairs of labeled sentences (including 549367 training, 9842 validation, and 9842 test samples). SNLI has a rich vocabulary and Keras' Tokenizer can identify 42391 unique words in its training set (note that we do not perform text preprocessing such as stemming or stop-word removal, which can reduce the size of vocabulary). 

\begin{figure}[h]
\begin{center}
   \includegraphics[width=0.8\linewidth]{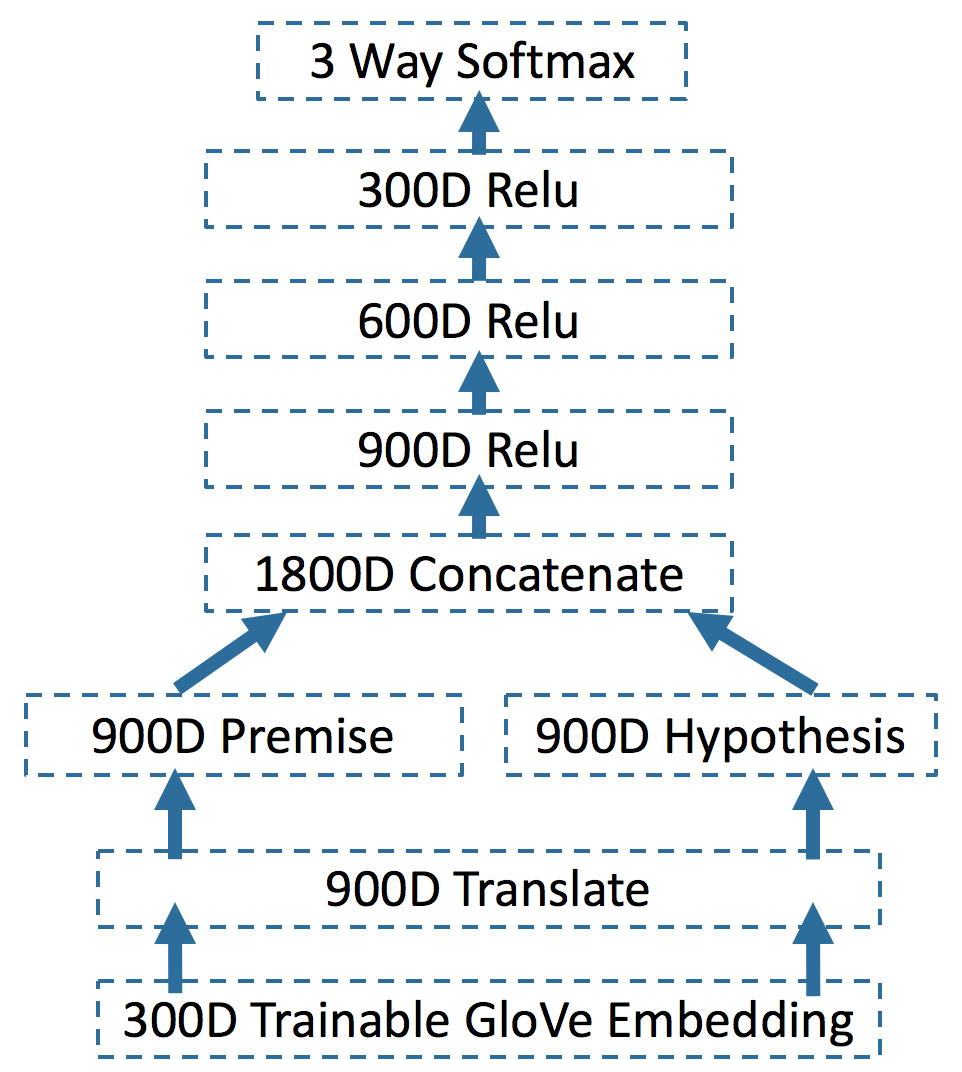}
\end{center}
   \caption{Design-I for SNLI}
\label{fig:desingI}
\end{figure}

We used SNLI dataset as the starting point for developing our model. Design-I for SNLI entails a 900D translate (Time Distributed) layer followed by 3 ReLU dense layers with decreasing size (900D-300D). The GloVe embedding layer for this design was set as $trainable= True$, which comparing with fixed embedding, increased the accuracy. Design-I has 15 millions total parameters, out of which 12 millions are assigned to the embedding layer, and only 7.2k of the parameters are non-trainable. There are also dropout of 0.5 and batch normalization layers added after dense layers for fine-tuning. Figure~\ref{fig:desingI} illustrates the outline of Design-I.  

Although the accuracy of the model on SNLI is fairly high (81\% on validation set), the result for modeling privacy policies was not satisfactory. Therefore, we decided to develop a model with a different dataset, which has a better performance for application to legal domain. We discuss the detailed results of Design-I in Section~\ref{evaluation}. 

\subsection{Design-II for MNLI}
Multi-Genre NLI Corpus or MNLI~\cite{nangia2017repeval} is a smaller dataset than SNLI, but its diversity of sentence forms, topics, and vocabulary (100157 words) is higher. MNLI consists of 393k training examples from five geners and 40k development and test examples. Due to this diversity, the network based on Design-I could not result in a satisfactory accuracy and thus we tried Design-II.

\begin{figure}[h]
\begin{center}
   \includegraphics[width=0.9\linewidth]{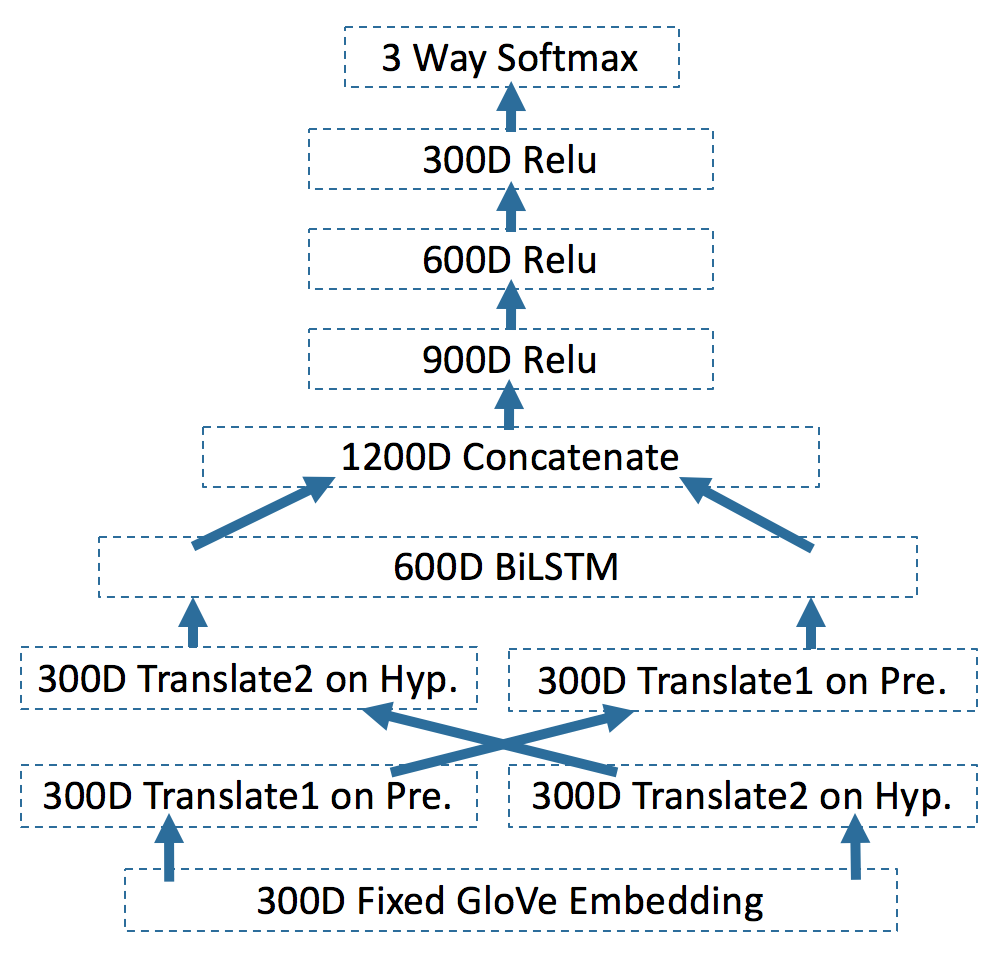}
\end{center}
   \caption{Design-II for MNLI}
\label{fig:desingII}
\end{figure}

In this design, we keep the GloVe embedding layer fixed, which reduces the number of trainable parameters (30 million non-trainable versus 3 million trainable). After embedding, there are two 300D translate layers that perform the intra- (Translate1) and inter-attention (Translate2) mechanism on the premise and hypothesis. Next, there is a BiLSTM layer followed by 3 ReLU layers with decreasing size (900D-300D). Figure~\ref{fig:desingII} shows the architecture of Design-II. 

\section{Evaluation and Discussion} \label{evaluation}

In this section, we report the results of training our models. First, we look at the data related to the privacy policies and then discuss the results of our models applied on this privacy policies. 

\subsection{Privacy Policies Dataset}
A privacy policy is an organizational document that describes what the data practices with respect to consumers' personal data are. Any online services, websites, apps, etc. are obliged by the laws to publish their privacy policies. PlayDrone~\cite{viennot2014measurement} is a public dataset that contains meta-data of over 1 million apps available on the Google Play store. Such meta-data also contain links to the apps' privacy policy pages. We created a crawler that can download and refine the data gathered from PlayDrone. Next, we perform a search on the collected privacy policies to find sensitive sentences that relate to the regulations in question. 

In this work, we focused on few rules from the General Data Protection Regulations (GDPR) to check the compliance of the privacy policies. 

\subsection{Results on SNLI}\label{ressnli}

\begin{figure}[h]
\begin{center}
   \includegraphics[width=0.9\linewidth]{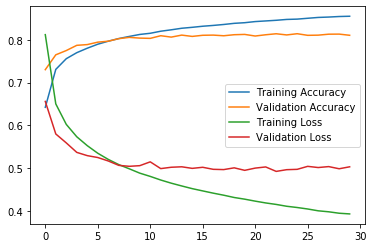}
\end{center}
   \caption{Accuracy graph for Design-I on SNLI}
\label{fig:snliaccuracy}
\end{figure}

Using Design-I, we trained the model with 30 epochs on SNLI dataset (batch size= 512), which resulted in 85\% training accuracy and 81\% validation accuracy. Test accuracy was also 80\%. Figure~\ref{fig:snliaccuracy} shows the graph for accuracy and loss on SNLI.

Although the accuracy of the model is fairly good, the model is not generalizable to be applicable on other geners, e.g. legal documents such as privacy policies. Here is a example of predictions for a test case.

\begin{quotation}
\noindent
    \textbf{Premise:} ``The data subject shall have the right to withdraw his or her consent at any time.'' (from GDPR article 7)\\
    \textbf{Hypothesis:} ``Test case one, you have no right to withdraw your consent.''\\ 
    \textbf{Predicted:} 0.41 contradiction, 0.37 neutral, 0.22 entailment.    
\end{quotation}

Despite the correct classification of the test case as a contradiction, the probability is very low (only 41\%).

\subsection{Results on MNLI}\label{resmnli}

\begin{figure}[h]
\begin{center}
   \includegraphics[width=0.9\linewidth]{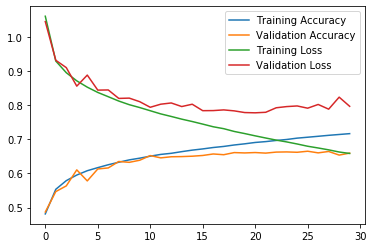}
\end{center}
   \caption{Accuracy graph for Design-II on MNLI}
\label{fig:mnliaccuracy}
\end{figure}

We started training the Design-II network without regularization techniques, which after 50 epochs resulted in 98\% training accuracy and 63\% validation accuracy. However, to resolve the overfitting and to increase the validation accuracy, we added dropout (0.2) layers to the network. A training of 25 epochs long with batch size of 1024, on 392702 training samples and 9815 validation examples, resulted in 70\% training accuracy and 66\% validation accuracy. Figure~\ref{fig:mnliaccuracy} shows the accuracy graph for Design-II on MNLI.

\begin{figure}[h]
\begin{center}
   \includegraphics[width=0.9\linewidth]{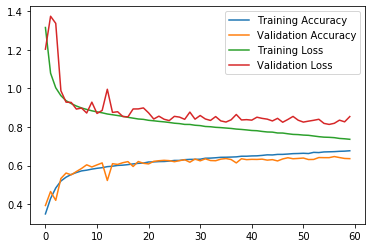}
\end{center}
   \caption{Accuracy graph for Design-II on Government Gener}
\label{fig:gmnliaccuracy}
\end{figure}

We also performed the training only on the government genre of MNLI (with 77350 training examples and 1945 validation samples) for 60 epochs. Although this training set has a smaller sample and vocabulary size (32928 words), its accuracy for entailment prediction on privacy policies is higher. With this model, we performed the following test cases (compare Hypothesis1 with Section~\ref{ressnli}): 

\begin{quotation}
\noindent
    \textbf{Premise:} ``The data subject shall have the right to withdraw his or her consent at any time.'' (from GDPR article 7)\\
    \textbf{Hypothesis1:} ``Test case one, you have no right to withdraw your consent.''\\ 
    \textbf{Predicted:} 0.99 contradiction, 0.01 neutral, and 0.0 entailment.\\
    \textbf{Hypothesis2:} ``Test case two, at any time you can withdraw your consent.''\\ 
    \textbf{Predicted:} 0.29 contradiction, 0.12 neutral, and 0.59 entailment.\\
    \textbf{Hypothesis1:} ``Test case three, You can not withdraw your consent any time.''\\ 
    \textbf{Predicted:} 1.0 contradiction, 0.0 neutral, and 0.0 entailment.
\end{quotation}
The accuracy graph for this model is also depicted in Figure~\ref{fig:gmnliaccuracy}.

Table~\ref{table1} shows a summary of the accuracy results for both designs on the datasets. Note that the government gener has samples only in the training and validation sets and not in the test set (thus N/A). 

\begin{table}[h]\label{table1}\begin{center}
\begin{tabular}{|c|p{1.75cm}|p{1cm}|p{1cm}|p{1cm}|}
\hline
Design & Dataset & Train Acc. & Val. Acc. & Test Acc. \\
\hline\hline
Design-I & SNLI & 85\% & 81\% & 80\% \\
Design-II & MNLI & 70\% & 66\% & 65\% \\
Design-II & Government & 67\% & 64\% & N/A \\
\hline
\end{tabular}\end{center}\caption{Summary of Results}
\end{table}

With respect to the results described here, we can see that besides the network design, the richness of a training dataset has a major role in improving NLI predictions. For example, a model trained with government literature is more adaptable for compliance checking on privacy policies than the one trained with SNLI. However, the main target for a dataset such as MNLI is to train a model that can be reliable for different, or even unseen, geners. For this reason, we supported our model with BiLSTM layers that can improve the accuracy of the predictions with respect to the sequential feature of textual data. For the semantic aspects, we used GloVe pre-trained embedding, which enhanced efficiency of our model in achieving high accuracy during training. However, other attention mechanisms such as a knowledge based approach~\cite{zhang2019knowledge} can help improving our work.  

\section{Conclusion}

We have shown our approach for analyzing privacy policies and their compliance with privacy regulations. We use an natural language inference model incorporating GloVe pre-trained word embedding and our attention mechanism uses BiLSTMs. Our approach is still in its preliminary stage, however, with the results achieved so far it seems fairly promising. The model trained on a specific gener of MNLI performs very well for finding inconsistencies in our test cases and other privacy policy statements. An improvement of our work can help automating detection of non-compliance in the privacy policies easier, which accordingly helps users to protect their data privacy. 

For the future work, we aim to enhance the attention mechanism of our work with a legal knowledge base and other network architecture features proposed in the state-of-the-are of NLI. For a better assessment of our work, we also aim to provide a labeled dataset of privacy policy statements, which facilitates evaluation of our model. Creating a more reliable dataset of privacy statement would also need a more robust search mechanism to find the relevant sections of privacy policies (other than ``user consent''). For this goal, we will use text matching and text similarity techniques to have a targeted choice of the privacy policy statements.

{\small
\bibliographystyle{ieee_fullname}
\bibliography{egpaper_final}
}

\end{document}